\documentclass[letterpaper]{article}
\usepackage{aaai}
\usepackage{times}
\usepackage{helvet}
\usepackage{courier}
\usepackage{algorithmic}
\usepackage{color}
\usepackage{subfig}
\usepackage{amsmath,graphicx}
\usepackage{amssymb}
\usepackage{algorithmic}
\author{Alireza Ghasemi$^1$ \\\\
	{\tt alireza.ghasemi@epfl.ch}\\\And
   Hamid R. Rabiee$^2$\\\\
   	{\tt rabiee@sharif.edu}\\\\
   	$^1$School of Computer and Communication Sciences\\
       Ecole Polytechnique F\'{e}d\'{e}rale de Lausanne (EPFL) \\
       Lausanne, Switzerland\\\And
    Mohammad T. Manzuri$^2$\\\\
        	{\tt manzuri@sharif.edu}\\\And
    M. H. Rohban$^2$\\\\
    	{\tt rahban@ce.sharif.edu}\\\\
    $^2$AICTC Research Center \\
    Sharif University of Technology \\
    Azadi Avenue, Tehran, Iran \\
    }
\DeclareMathOperator*{\argmin}{arg\,min}

\begin{document}
%
\title{A Bayesian Approach to the Data Description Problem}
\maketitle
\begin{abstract}
\begin{quote}
In this paper, we address the problem of data description using a Bayesian framework. The goal of data description is to draw a boundary around objects of a certain class of interest to discriminate that class from the rest of the feature space. Data description is also known as one-class learning and has a wide range of applications.

The proposed approach uses a Bayesian framework to precisely compute the class boundary and therefore can utilize domain information in form of prior knowledge in the framework. It can also operate in the kernel space and therefore recognize arbitrary boundary shapes. Moreover, the proposed method can utilize unlabeled data in order to improve accuracy of discrimination.

We evaluate our method using various real-world datasets and compare it with other state of the art approaches of data description. Experiments show promising results and improved performance over other data description and one-class learning algorithms.

\end{quote}
\end{abstract}

\section{Introduction}

A critical assumption for many supervised learning algorithms is presence of training data from all classes under study. It means that, for example a binary classification algorithm requires training samples of both classes in order to work properly. In scenarios where this condition is not met, performance degrades considerably or even algorithm fails to run. A well-known example of such scenarios is the problem of image retrieval (with relevance feedback) in which the system is given only rare samples of the relevant class and therefore traditional supervised learning algorithms are not suitable for this problem. 

The aforementioned problems are known as data description or one-class learning problems and have a wide range of applications from pattern recognition to data mining and image processing. Information retrieval, video surveillance, outlier detection and object detection are all among applications of one-class learning algorithms.

As well as presence of samples of only one class (which is called target class), there are also other scenarios in which one-class learning can be beneficial. Another implicit assumption of many supervised learning algorithms is that the prior probabilities of different classes in the training set (and whole feature space) are equal or at least very close. However, this is also violated in many real-word situations such as spam detection in which the proportion of spam messages and regular e-mail is quite different in a fair data sample. One-class learning algorithms can also be beneficial in this case since they do not assume this and are designed for databases in which the proportion or other properties of different classes (like statistical distribution) are quite different. Examples of other problems of this kind are industrial fault detection and information retrieval.

Several one-class learning algorithms have been proposed so far. The work in \cite{ocls} is a recent survey on current trends in one-class learning. Many of these algorithms are extensions of traditional classification algorithms adapted to work in one-class settings. For example, in \cite{bishop2} an approach based on neural networks is proposed for novelty detection. Also in \cite{ocdt} a variant of decision tree has been used for one-class learning. In \cite{ocknn1} the $k$ nearest neighbors algorithm has been used for one-class learning. Although such algorithms are simple and easy to understand, they are usually inefficient on complicated real-world data.

A major class of one-class learning algorithms are based on statistical density estimation. These approaches assume a parametric statistical model for the target class and then estimate the parameters of that model. The likelihood of a data sample measures the degree that the sample belongs to the target class. In \cite{kde1} approaches based on Parzen or kernel density estimation have been proposed. Also in \cite{gmm}, Gaussian mixture models have been utilized for novelty detection. The principal advantage of these methods is the rigid theoretical foundations behind them. However, they can not directly operate in the kernel space and therefore have some limitations in modeling the complex boundary shapes.

Since the introduction of support vector machines \cite{svm} and kernel methods \cite{shawe}, there has been a growing interest in adapting  kernel-based approaches to one-class learning. Scholkopf in \cite{ocsvm} presented one-class SVM. It is a variation of traditional binary SVM which tries to separate target data from the axis origin. \cite{svdd} proposed support vector data description. In this method, a hypersphere of minimum volume is sought to surround the target samples. In \cite{asdd}, it is shown that the two approaches yield the same solution when the used kernel is isotropic. Kernel methods yield good results in most problem and model different kinds of boundary shapes utilizing flexibility of the kernel space. However, domain knowledge can not be easily embedded into kernel methods. Moreover, these methods can not directly utilize unlabeled data to improve accuracy.

Utilizing unlabeled data in the process of one-class learning has also been of interest in recent years. \cite{lfpu},\cite{mapcon} and \cite{lfpso} have utilized unlabeled samples as well as positive target samples in the process of one-class learning.  These methods try to infer a set of confident negative samples from the unlabeled set and then perform a traditional binary classification algorithm. \cite{asdd} and \cite{svdd} have utilized outlier samples in addition to targets in the process of learning. The relation between support vector methods and density based approaches has been discussed in \cite{precise}. In \cite{induced}, the local density around target point has been used to improve the SVDD.

The Gaussian process regression method has been adapted for one-class learning in \cite{ocgp}. Moreover, among the probabilistic approaches to one class learning, \cite{bod} has used a Bayesian approach which defines a regression model over data samples. In \cite{bod2,bod4} Bayesian approaches have been used for outlier detection. In \cite{bod3} a Bayesian approach has been used to detect outliers in ordinal data. These methods are more flexible since they allow uncertainty in the model and use domain knowledge in constructing the classifier. However, their principal drawback is their computational inefficiency and lack of sparseness.

In this paper, we propose a novel Bayesian approach to the data description problem. The principal contribution of our work is twofold: First we develop a Bayesian framework which can benefit from advantages of both probabilistic and support vector approaches. For example our approach can generate sparse solutions and in addition, we propose a principled method for utilizing prior knowledge in the process of one-class learning. The second contribution of our work is that the proposed approach can benefit from unlabelled data in improving the accuracy of classification.

In the rest of this paper, after reviewing SVDD, a well-known data description algorithm, we propose our approach and study its properties and extensions. Then, we thoroughly test our approach against other one-class learning algorithms under various conditions.


%
%
\section{The Traditional Support Vector Data Description}
\label{svddint}

Support vector data description is a well-known algorithm for one-class learning which has been widely used in various applications. It is a kernel-based approach which tries to find a hypersphere which is as small as possible and meanwhile contains as much target data as possible, hereby avoiding outlier samples. This goal is achieved by solving a convex optimization problem over the target data points in the kernel space, in a way very similar to the well-known support vector machine algorithm.

We describe SVDD briefly in the rest of this section. For a more detailed explanation, refer to the seminal work of Tax and Duin. 

%

Suppose we are given a dataset $\{x_1,\ldots, x_n\}$ which consists of the training set. The main idea of support vector data description is to find a hypersphere in the feature space containing as many of the training samples as possible  while having minimum possible volume. To achieve this goal, data are first transformed to a higher dimensional kernel space in which support of the data is a hypersphere.

The sphere is characterized by its center $C$ and radius $R > 0$. The minimization of the sphere volume is achieved by minimizing its square radius $R^2$. Data samples outside the hypersphere are penalized in the objective function. To consider the penalty, slack variables $\xi_i \geq 0$ are introduced and the optimization problem is formulated as:

\begin{equation}
\mathrm{min}_{R\in \mathcal{R} , \xi_i\in\mathcal{R}^n, C\in\mathcal{F}}\;R^2+\frac{1}{N\nu}\sum_{i=1}^{N} \xi_i
\end{equation}
such that
\begin{equation}
||\phi(x_i)-C|| \leq R^2+\xi_i \;\mathrm{ and }\; \xi_i \geq 0
\end{equation}.

In the above formula, $\phi$ is the transformation which maps data points to the higher dimensional. The parameter $\nu$ controls the trade-off between the hypersphere volume and the proportion of samples in the hypersphere. It can also be used to control the sparseness of the solution of the optimization problem \cite{asdd}.

Introducing Lagrangian multipliers to account for constraints, we obtain the following dual problem:

\begin{equation}
\label{svdual}
\mathrm{min}_{\alpha} \;\alpha^t\mathbf{K}\alpha - \alpha^tdiag(\mathbf{K}) 
\end{equation}
such that
\begin{equation}
0 \leq \alpha_i \leq \frac{1}{N\nu} \;\mathrm{and}\; \sum \alpha_i=1
\end{equation}

In \eqref{svdual}, $\mathbf{K}$ is the kernel matrix in which $\mathbf{K}_{i,j}=<\phi(x_i),\phi(x_j)>$ and $diag(\mathbf{K}) $ is the main diagonal of $\mathbf{K}$. One may notice that it is not needed to explicitly transform data to the kernel space and only defining a kernel function (dot product between transformed data) in terms of original points is sufficient. We call this function $\mathcal{K}(.,.)$. Therefore, $\mathcal{K}(x_i,x_j)=<\phi(x_i),\phi(x_j)>$.

Solving the dual optimization problem yields vector $\alpha$ in which most of the values are 0. Samples $x_i$ with positive $\alpha_i$ are called support vectors of the SVDD. center $C$ of the hypersphere can be specified in term of Lagrange multipliers $\alpha_i$ as:

\begin{equation}
C=\Sigma_i \alpha_i \phi(x_i)
\end{equation}.

We can rank test samples by their proximity to the center of the hyper sphere. The ranking function $f$ is defined as below in which smaller values of $f$ mean more similarity to the target class.
\begin{center}
\begin{equation}
\begin{split}
f(z)=\Sigma_i\Sigma_j \alpha_i\alpha_j \mathcal{K}(x_i,x_j) + \mathcal{K}(z,z) -2\Sigma_i \alpha_i\mathcal{K}(x_i,z)
\end{split}
\end{equation}
\end{center}

\section{The Bayesian Approach}
\label{bddint}
As we saw in the previous section, the support vector data description algorithm finally reduces to finding center of the surrounding hypersphere in the embedded space as a weighted average of sample target points in which many of the weights are zero.Data points for which the corresponding weight is non-zeros are called support vectors.

In this section we derive the proposed Bayesian data description method. We look at the problem of data description from a different point of view. Later we show that the interpretation of data and parameters in our model is equivalent to that of SVDD. 

Our method is based on the same set of parameters as the SVDD (in its dual form), i.e we will try to find a vector of weights, one for each data sample. Assume that we transform all data samples using the mapping $\phi$ to a higher dimensional embedded (kernel) space in which transformed data follow a Gaussian distribution with covariance matrix  $I$ and mean $\Sigma_i \alpha_i \phi(x_i)$. i.e.:

\begin{equation}
\label{wgauss}
\phi(x_j) \sim \mathcal{N}(\Sigma_i \alpha_i \phi(x_i),I)
\end{equation}

We limit $\alpha_i$ values to form a convex set, i.e. $0<\alpha_i<1$ and $\Sigma_i \alpha_i=1$. Later, we will discuss the reason behind this assumption.

The main difference between the estimation in \eqref{wgauss} and the conventional Gaussian density estimation is that the mean is limited to be a weighted average of training target points. Hereafter, we call this model the weighted Gaussian model.

The principal correspondence between the weighted Gaussian model and the SVDD is that the mean of the weighted Gaussian is equivalent to the center of the hypersphere in the SVDD. Therefore, distance of a point to center of the surrounding hypersphere in the SVDD model is inversely proportional to the likelihood of a data point in the weighted Gaussian model. We use this fact to show that SVDD is itself a special case of the weighted Gaussian model. Then we improve upon SVDD equivalent case of weighted Gaussian by utilizing unlabeled data and defining more precise prior knowledge.

To achieve this goal, first we have to estimate parameters of the weighted Gaussian model using a statistical parameter estimation approach. Various parameter estimation methods have been proposed in the literature so far. Two of the most common ones are maximum likelihood ones and Bayesian approach.

Maximum likelihood estimation is a simple optimization-based approach which maximizes the likelihood of training data with regard to the unknown parameters. However, this method is not flexible and can not utilize domain information to improve the estimation. We seek to arbitrarily constrain the estimation procedure toward finding solution with specific properties (e.g. sparseness) and moreover utilize various forms of domain information in this procedure. Therefore, we use the Bayesian estimation.

In Bayesian parameter estimation, a prior distribution $p(\alpha)$ is defined over parameter vector $\alpha$ and the posterior probability $p(\alpha|D)$ is derived by applying the Bayes rule:

\begin{center}
\begin{equation}
\begin{split}
p(\alpha|D)=\frac{p(D|\alpha)p(\alpha)}{p(D)}
\end{split}
\end{equation}
\end{center}

In which $p(D|\alpha)$ is the likelihood of training data given a specific value of  $\alpha$ and $p(D)$ is a normalizing constant.

We assume that the parameter vector $\alpha$ follows a Gaussian distribution with mean $m$ and covariance matrix $C$ i.e.:

\begin{equation}
\alpha \sim \mathcal{N}(m,C)
\end{equation}

Applying the Bayes rule, we have:

\begin{equation}
\label{brule}
p(\alpha|D) \propto p(D|\alpha)p(\alpha)
\end{equation}

We have omitted $p(D)$ in \eqref{brule} because it is independent of $\alpha$.

Maximizing the a posteriori probability of $\alpha$ (MAP estimation) we will have:

\begin{center}
\begin{equation}
\label{bddopt}
\begin{split}
\hat{\alpha}=\argmin_{\alpha} \;\alpha^t(n\mathbf{K}+C^{-1})\alpha-2\alpha^t(\mathbf{D}\mathbf{1}+C^{-1}m)
\end{split}
\end{equation}
\end{center}

Matrix $\mathbf{D}$ is the diagonal matrix  of weighted degree of samples, i.e. $\mathbf{D}_{i,i}=\Sigma_j \mathbf{K}_{i,j}$.

Equation \eqref{bddopt} is the key to our approach since it allows purposeful modification of the behavior of the final solution by setting different values for covariance matrix $C$ and mean $m$ of the parameter vector. For example, the objective function of SVDD can be derived from \eqref{bddopt} by choosing the appropriate $C$ and $m$ (We can check this by substituting $C=I$ and $m=diag(\mathbf{K})-\mathbf{D}\mathbf{1}$ and assuming an isotropic Kernel ). Moreover, wee see that the optimization only depends on dot products of points in the embedded (kernel) space. Therefore, the Bayesian estimation to the weighted Gaussian model is itself a kernel method. That is why we constrained the mean of the model to be a weighted average of training points.

The most trivial choice for the parameters could be setting $C$ to the identity matrix and each $m_i$ equally to $\frac{1}{n}$. However, this kind of prior knowledge is non-informative and therefore yields the same non-sparse trivial solution as the maximum likelihood approach. 

As a better and more informative prior knowledge, we could modify the mean vector $m$ such that the data points which lie in a dense area of the embedded space receive smaller prior weight. The main motivation behind this choice is the fact that target points located in dense areas of feature space are less likely to be close to boundary of the target class and therefore their corresponding weight should have more prior probability of being close to zero. In contrast, target points located in less-dense areas of feature space are more likely to be on or close to the boundary and therefore their corresponding weight should be a priori larger than other points.

With these facts in mind, we suggest that the mean of prior probability of parameter vector be proportional to $–\mathbf{D1}$, in which $\mathbf{D}$ is the same diagonal matrix as in \eqref{bddopt}. This is reasonable because the weighted degree of a point is a good approximation of local density of the area near that point. Therefore we set:

\begin{center}
\begin{equation}
\label{mlf}
\begin{split}
m_i\propto-\Sigma_{j\in L} \mathbf{K}_{i,j}
\end{split}
\end{equation}
\end{center}

for each element of mean vector $m$. Using such prior knowledge, we expect that samples crucial in determining center of the Gaussian become much more likely to receive larger values. This causes the solution to become sparse and more accurately capture the underlying distribution and its support (boundary).

The pseudo code for the Bayesian data description algorithm is depicted in figure \ref{bddalg}. In this algorithm, parameter $0<\nu<1$ controls sparsity of the solution.  Larger values for $\nu$ cause the solution to become as sparse as possible. $\nu$ can also be used to reduce the effect of outlier data on the final solution.

\begin{figure}
\tiny
\begin{algorithmic}[1]

\REQUIRE Set of Target Training Samples $T$
\REQUIRE  Set of Target Test Samples $S$

\STATE Compute Kernel matrix $\mathbf{K}$ from training data
\STATE Compute diagonal matrix $D$ such that $D_{ii}=\Sigma_j \mathbf{K}_{i,j}$
\STATE $n\leftarrow$ \emph{number of training samples}
\STATE $n_{test}\leftarrow$ \emph{number of test samples}
\STATE  $C\leftarrow I_{n\times n} $
\STATE $\forall i:  1\leq i \leq n \rightarrow m_i=-(\Sigma_j \mathbf{K}_{i,j})^\nu$
\STATE $\alpha = \argmin_{\alpha} \;\alpha^t(n\mathbf{K}+C^{-1})\alpha-2\alpha^t(\mathbf{D}\mathbf{1}+C^{-1}m)$
\FOR{$i = 1 \to n_{test}$} 
\STATE $score_i=\Sigma_j\Sigma_k \alpha_j\alpha_k \mathcal{K}(x_j,x_k) + \mathcal{K}(x_i,x_i)-2\Sigma_j \mathcal{K}(x_i,x_j)$
\ENDFOR
\STATE Sort test samples in ascending order by the $score$ values
\RETURN Desired number of samples from top of the list

\end{algorithmic}

\caption{Bayesian data description (BDD) algorithm}
\label{bddalg}
\end{figure}
\begin{figure*} [!t]
\centerline{
\subfloat[SVDD]{\includegraphics[width=0.5in]{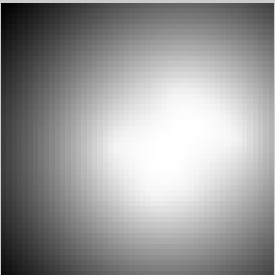}%
\label{sdatasvdd}}
\hfil
\subfloat[ML]{\includegraphics[width=0.5in]{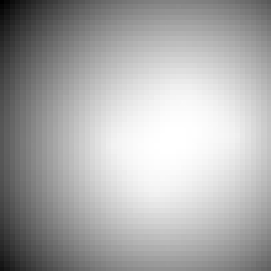}%
\label{sdatakde}}
\hfil
\subfloat[Bayesian]{\includegraphics[width=0.5in]{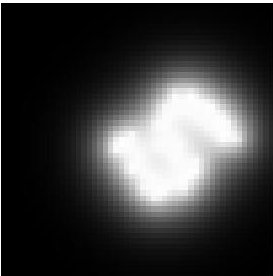}%
\label{sdatagmm}}
}
\caption{Performance of density estimation for SVDD against the proposed weighted Gaussian}
\label{bdistribs}
\end{figure*}

Also figure \ref{bdistribs} depicts performance of our weighted Gaussian model (with maximum likelihood and Bayesian estimation) in capturing a typical S-shaped distribution and compares it with that of  SVDD. We see that weighted Gaussian with maximum likelihood estimation has captured a mostly spherical distribution shape which shows that this method lacks sparsity and flexibility and its solution is close to the simple mean of points which is the trivial solution. Both BDD and SVDD has been more successful in capturing the real shape of the distribution and avoid over-fitting. 

\subsection{Utilizing Unlabelled Data}
\label{ssdd}

Methods which utilize unlabeled data to improve learning accuracy have received much attention in recent years. These methods use unlabeled data to infer information about geometry of data and the corresponding manifold. Such information can be used to improve the accuracy of supervised classifiers.

In the Bayesian data description approach, information about the geometry of data can be utilized to determine the prior probability distribution of the parameter vector . Since we use local density of area around points in determining the prior probability distribution of the model parameters. The information we gather from unlabeled samples can be useful in determining the local density around a point, more accurately. Having unlabeled data available, we can now set:
\begin{center}
\begin{equation}
\label{mlf}
\begin{split}
m_i=-\Sigma_{j\in L\cup U} \mathbf{K}_{i,j}
\end{split}
\end{equation}
\end{center}
In which $L$ and $U$ are the set of labeled and unlabeled data, respectively. 

Another parameter which can be modified by using unlabeled data is the covariance matrix $C$. Information about geometry of data can be used in constructing this matrix by using any type of data dependent kernel.

An example of using unlabeled data for adjusting the covariance is by computing the Laplacian operator of training samples \cite{surv1}.

Suppose we define a $k$-nn graph over all data samples. A $k$-nn graph is a graph in which nodes are data samples and each sample is connected to its $k$ nearest neighbors. Weight $W_{i,j}$ of each edge is proportional to the similarity between data samples $x_i$ and $x_j$. Gaussian function is a popular choice for $W$. 

Having the weight matrix, the Laplacian $L$ of the graph is defined as $L=D-W$ in which $D$ is a diagonal matrix in which $D_{i,i}=\Sigma_j W_{i,j}$. Utilizing the Laplacian, we adjust matrix $C$ as:
\begin{center}
\begin{equation}
\label{mlf2}
\begin{split}
C^{-1}=(L^{-1})_{1\dots n, 1\dots n}
\end{split}
\end{equation}
\end{center}
Utilizing unlabeled data in this way, manifold of all data (target and non-target) is modeled in the $C$ matrix, whereas manifold of target data can be modeled in the kernel matrix $\mathbf{K}$ of the weighted Gaussian itself. Therefore, we use both manifolds to better model distribution of the data.

\subsection{Time Complexity of the Bayesian Data Description}
Constructing the prior vector $m$ can be done at the time of constructing the kernel matrix and requires $\mathcal{O}(n^2)$, the same as minimum complexity of kernel construction (in the general sense). The objective function of the BDD is a convex quadratic programming problem which can be solved in $\mathcal{O}(n^3)$ time. SVDD also reduces to a quadratic programming problem. Therefore the time complexity of BDD is not higher than SVDD.

In the semi-supervised settings (SSDD), we require to compute inverse of the covariance matrix which is of complexity $\mathcal{O}((n+m)^3)$ ($m$ is the number of unlabeled samples). The prior weight vector can still be formed at the time of kernel construction with the same complexity as kernel construction ($\mathcal{O}((n+m)^2)$). Finally, the resulting quadratic requires $\mathcal{O}(n^3)$ time to be solved which is independent of the number of unlabeled data.
\section{Experimental Results}
\label{exps}

\subsection{Experiment Setup}

Various datasets from the UCI repository \cite{uci}, as well as Corel \cite{corel} and Caltech-101 \cite{caltech} image databases were used for experiments. Their properties are depicted in table \ref{datasets}. 

\begin{table}[!t]
\tiny
\renewcommand{\arraystretch}{1.3}
\caption{Datasets Used in Experiments}
\label{datasets}
\centering
\begin{tabular}{|l|c|c|c|}
\hline
{\bf Dataset} & {\bf No. of features} & {\bf No. of Samples} & {\bf  No. of Classes}\\ \hline

{\bf Iris }& 256 & 150 &  3 \\ \hline

{\bf USPS}& 256 & 9998 &  10 \\ \hline

{\bf Pendigits}& 16 & 10992 &  10\\ \hline

{\bf ISOLET}& 617 & 7797 & 10\\ \hline

{\bf MNIST }& 784 & 60000 & 10\\ \hline

{\bf COIL20 }& 1024 & 1440 & 20\\ \hline

{\bf Caltech-101} & 144 & 9144 & 101 \\ \hline

{\bf Letter} & 16 & 1259 & 26 \\ \hline
{\bf Glass} & 9 & 214 & 6 \\ \hline
{\bf Corel} & 144 & 1000 & 20 \\ \hline

\end{tabular}
\end{table}

In each experiment, one of the classes was selected as target, and all other samples were treated as outlier. One-half of the target samples were selected for training. The rest of training samples, as well as outlier data were selected as test samples. For the Caltech-101 and Corel image datasets, feature extraction was performed by the CEDD \cite{cedd} feature extraction algorithm.

SVDD method and one-class Gaussian process were implemented and compared with the proposed BDD method. The Gaussian function was used as the kernel. The parameters of the classifiers and the kernel were adjusted by 10-fold cross validation. All sample selections were done by random sampling.

For measuring efficiency of one-class learning, we computed precision in the top $k$ returned results as accuracy measure and set $k$ to the (estimated) number of target samples in the test set. This measure has the advantage that unlike precision or recall, we don't need to compute more than one quantity in order to achieve meaningful results. Moreover, the value chosen for k eliminates the need for explicitly adjusting an acceptance threshold for one-class learning algorithms which could be a tedious task and have significant effect on functionality of algorithms.

\subsection{Experiments}

In table \ref{dsbdd} we compare the performance of BDD with that of SVDD and one-class Gaussian process. 
The BDD and SVDD show similar performance with slight improvements in BDD because of utilizing the density-based prior knowledge. One-class Gaussian process also has a reasonable performance but this algorithm is not sparse and therefore lacks benefits of the other models and is more hardly generalizable. Running times (in seconds) of algorithms are depicted in parentheses in each cell.

\begin{figure*}[!t]
\centering
\includegraphics[width=4.5in,height=0.6in]{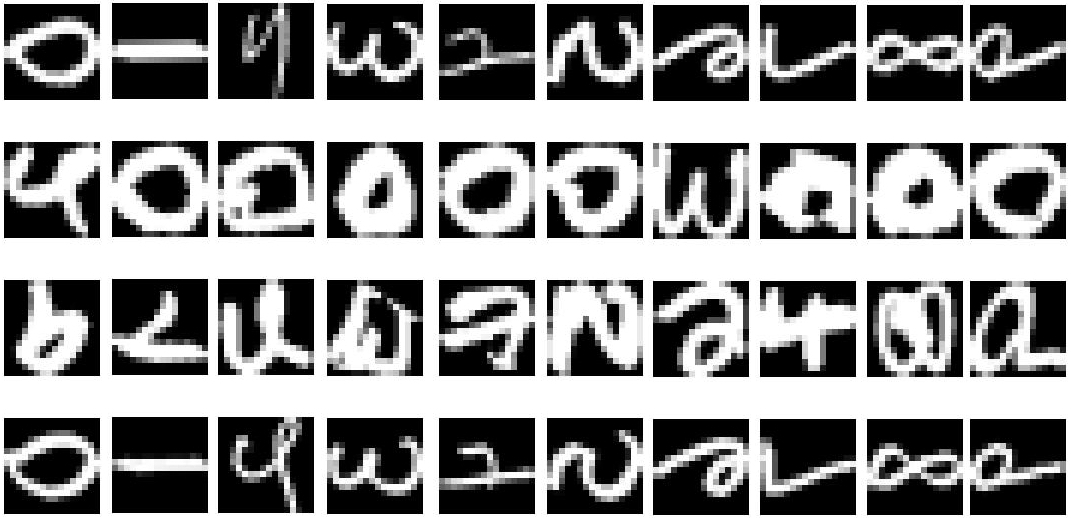}

\caption{ Performance of BDD on different classes of the USPS dataset}
\label{uspsall}
\end{figure*}

Figure \ref{uspsall} shows interesting results about performance of the Bayesian data description on different classes of the USPS digit recognition dataset. Here, we visualize different data samples in order to understand the operation of BDD. Each column depicts performance on one class of the USPS dataset. 

The firs row shows the most likely samples of each class returned by the BDD algorithm. As can be seen, all samples in this row have been classified correctly and are appropriate representatives for their respective class. 

The second row shows the least likely sample detected as target by the BDD for each class. We can see that these samples are misclassified data and count as error rate of the classifier. It is reasonable to have error here since we rank data samples by likelihood to the target class and samples with lower ranks are more likely to be misclassified (unless the precision is perfect 1).

The third and fourth row deal with the prior estimation of the local density around each sample which is done by computing its weighted degree. The third row shows the data sample with least weighted degree. We see that these samples usually can not be considered typical representatives of their underlying target class and should be far from the center of mass of the target class. These data samples lie in the boundary of target class and therefore have the most important role in defining the center of the weighted Gaussian model. Because of this property of weighted degree of data samples, we set the prior probability of the parameter corresponding to each sample proportional to the weighted degree of that sample.

The fourth row shows the sample with largest weighted degree. We can see that the data samples are typical representatives of their underlying class. This is because samples with large weighted degree usually lie within the target hypersphere and are far from the boundary of the target class.

\begin{figure*}[!t]
\centerline{
\subfloat[Pendigits]{\includegraphics[width=3.3in,height=0.7in]{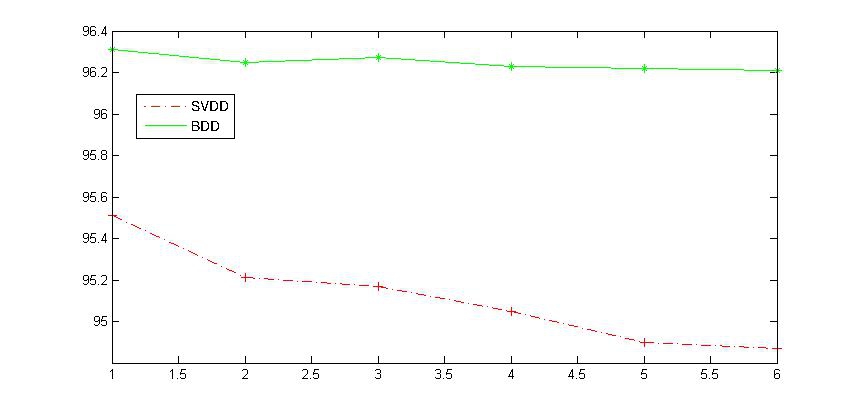}%
\label{uspso}}
\hfil
\subfloat[Corel]{\includegraphics[width=3.3in,height=0.7in]{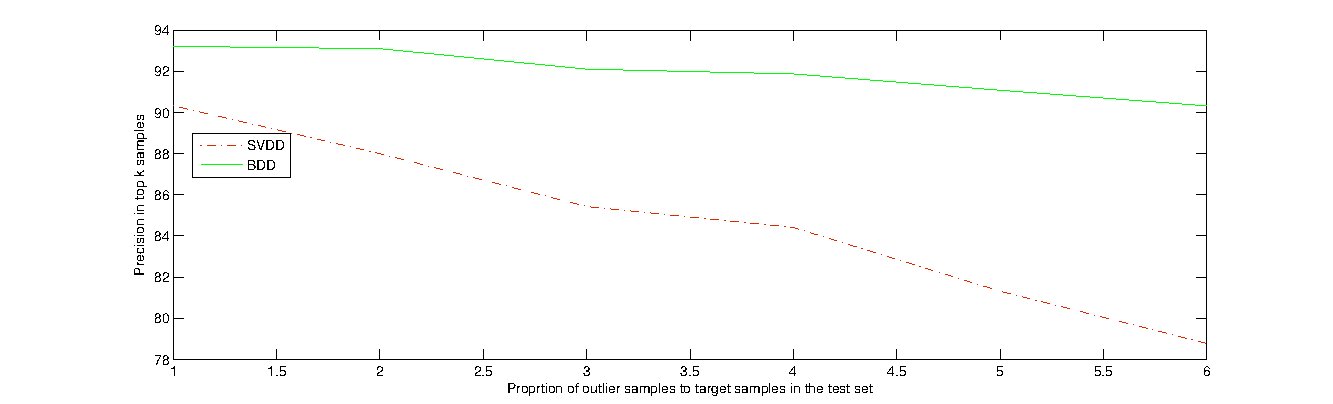}%
\label{corelo}}
}
\caption{Precision against proportion of target and outlier samples for big and small datasets}
\label{bigpo}
\end{figure*}

An important point to note about one-class learning algorithms is their sensitivity to the proportion of target and outlier data samples in the test set. The accuracy of the resulting model can be affected significantly by varying this ratio. We test this by gradually increasing the proportion of outlier samples in the test set and computing precision in each case. Figure \ref{bigpo} depicts results of studying this property for SVDD and BDD model. 

As can be seen in figure \ref{uspso}, in large datasets the precision of classification is not affected largely by increasing the proportion of outlier samples. This is mostly because the training set is big enough to capture the distribution of target class. Presence of sufficient target samples prevents probably noisy data to affect misclassification rate.

However, figure \ref{corelo} depicts that this is not the case for smaller datasets like Corel and Caltech-101. Here, due to insufficiency of target training samples, noisy data can significantly influence the  boundary of target class and hence misclassification rate increases by increasing the proportion of outlier samples.

We can see in both figures that BDD is less sensitive to variations in the proportion of outlier samples, which is mostly because of its use of prior knowledge over model parameters. By using weighted degree as a prior, we prevent noisy data to become significant in constructing the model and compensate for the model uncertainty.
%
%
%
%
%
%
%
%
\begin{table}[!t]
\renewcommand{\arraystretch}{1.3}
\caption{Experimental results with supervised Bayesian data description and other one-class learning algorithms}
\label{dsbdd}
\centering
\tiny
\begin{tabular}{|c|c|c|c|c|}
\hline
{\bf Dataset} & {\bf OCGP} & {\bf SVDD} &  {\bf  BDD}  \\ \hline

{\bf Iris }& $97.85\pm0.13 (0.82)$ &$98.08\pm0.08 (0.54)$ &   \textbf{98.11$\pm$0.03(0.35)} \\ \hline

{\bf USPS}& $89.22\pm0.04 (2.10)$ &$89.14\pm0.03 (1.29)$ &   \textbf{89.23$\pm$0.03(1.43)} \\ \hline

{\bf Pendigits}& $95.75\pm0.22 (1.92)$ &$94.65\pm0.10(1.61)$ &   \textbf{95.91$\pm$0.12(1.54)} \\ \hline

{\bf ISOLET}& $91.28\pm0.87 (2.31)$ &$92.37\pm0.51(1.21)$ &   \textbf{94.52$\pm$0.54(1.24)} \\ \hline

{\bf MNIST }& $87.46\pm0.92 (3.86)$ &$85.01\pm0.30(3.82)$ &   \textbf{88.51$\pm$0.34(3.95)} \\ \hline

{\bf COIL20 }& $51.33\pm3.04 (2.32)$ &  \textbf{58.54$\pm$1.87 (1.74)} &   $57.01\pm2.38(1.79)$ \\ \hline

{\bf Caltech-101}& $79.83\pm1.02 (2.14)$ &$80.07\pm0.91(2.18)$ &   \textbf{82.21$\pm$0.60(2.01)} \\ \hline

{\bf Letter}& \textbf{83.10$\pm$1.19 (1.14)} &$80.23\pm0.81(0.93)$ &   $82.41\pm0.94(0.91)$ \\ \hline
{\bf Glass} & $77.91\pm1.81 (0.14)$ &$77.12\pm1.70(0.09)$ &  \textbf{79.34$\pm$1.72(0.09)} \\ \hline
{\bf Corel}& $92.21\pm1.51 (1.58)$ &$90.16\pm1.17(1.41)$ &   \textbf{93.19$\pm$1.19(1.41)} \\ \hline
\end{tabular}
\end{table}


\subsubsection{Experiments with unlabeled data}

For semi-supervised learning,  we divided the training set into a labeled and an unlabeled set. We set the size of unlabeled set twice the size of labeled training set and for better runtime, used the unlabeled data only to improve the prior mean of Bayesian model. In addition to SVDD, the mapping-convergence algorithm \cite{mapcon} was also implemented and used in comparisons. Results of semi-supervised learning are depicted in figure \ref{datasets-ss}.

We see in table \ref{datasets-ss} that semi-supervised Bayesian data description algorithm (SSDD) outperforms other approaches. Since SVDD can not use unlabeled data, it is expectable that we don't see any performance improvement by adding unlabeled data. Mapping-convergence also does not achieve good performance, because this algorithm uses unlabeled data only to select some confident negative samples and then performs a traditional binary classification algorithm. Therefore, the problems that arise for binary classification on one-class problems also arises for this algorithm and degrades its performance. Moreover, we can see that smaller and more difficult datasets are improved more significantly by utilizing unlabeled data. This is because of the fact that the training data are insufficient for these problems and therefore they benefit more from using the unlabeled data.

Also running time (in seconds) of each algorithm is depicted in parentheses in table \ref{datasets-ss}. We can see that SSDD performs quite faster than mapping-convergence and also it's speed is very close to that of SVDD that does not use unlabeled data at all. The mapping-convergence algorithm is slower than SSDD because it runs both one-class learning (to detect negative points) and a traditional binary classification, whereas SSDD only runs data description.

\begin{table}[!t]
\renewcommand{\arraystretch}{1.3}
\caption{Experimental results with semi-supervised Bayesian data description learning algorithms.}
\label{datasets-ss}
\centering
\tiny
\begin{tabular}{|c|c|c|c|}
\hline
{\bf Dataset} & {\bf SVDD} & {\bf MC} & {\bf  SSDD}  \\ \hline

{\bf Iris }& $98.06\pm0.09 (0.52)$ &$98.17\pm0.04 (1.54)$ &  {\bf 99.89$\pm$0.01(0.75)} \\ \hline

{\bf USPS}& $89.19\pm0.04 (1.30)$ &$88.23\pm0.02 (2.57)$ &  {\bf 94.76$\pm$0.05(2.16)} \\ \hline

{\bf Pendigits}& $94.75\pm0.12 (1.72)$ &$96.01\pm0.07(2.70)$ &  {\bf 98.89$\pm$0.10(1.98)} \\ \hline

{\bf ISOLET}& $92.28\pm0.57 (1.22)$ & $94.87\pm0.23(2.60)$ &  {\bf 98.23$\pm$0.38(2.02)} \\ \hline

{\bf MNIST }& $85.06\pm0.32 (3.81)$ &$90.01\pm0.18(9.82)$ &  {\bf 94.48$\pm$0.31(5.07)} \\ \hline

{\bf COIL20 }& $54.63\pm2.00 (1.74)$ &$59.25\pm1.06 (4.46)$ &  {\bf 66.53$\pm$2.51(2.75)} \\ \hline

{\bf Caltech-101}& $80.01\pm0.86 (2.14)$ &$83.07\pm0.31(4.78)$ &   {\bf 89.90$\pm$0.58(3.21)} \\ \hline

{\bf Letter}& $80.20\pm0.91 (1.00)$ &$88.34\pm0.30(2.06)$ &   {\bf 95.10$\pm$0.79(1.61)} \\ \hline
{\bf Glass} & $77.20\pm1.71 (0.10)$ &$79.12\pm1.01(1.00)$ &  {\bf  86.02$\pm$1.61(0.86)} \\ \hline
{\bf Corel}& $90.21\pm1.21  (1.48)$ &$93.69\pm0.89(3.41)$ &   {\bf 97.26$\pm$1.24(2.52)} \\ \hline
\end{tabular}
\end{table}

\section{Conclusions}

In this paper, we proposed a novel Bayesian approach for the data description problem which has various applications in machine learning. Our approach is a bridge between probabilistic and kernel based data description and hence can use benefits of both types of approaches such as sparseness of the support vector approaches and utilizing prior knowledge in the probabilistic approaches. Moreover, our approach can utilize unlabeled data in order to improve accuracy of the data description.

The prior knowledge utilized in our model can have various applications. For example, the information in the data samples prior, can be used to estimate most probable support vectors and reduce the size of data set, hereby reducing time complexity of the training. Moreover, robustness of the algorithm to noise could be further improved.


\section*{Acknowledgment}

The authors thank the AICTC research center and VAS Laboratory of Sharif University of Technology.

\bibliographystyle{aaai}

\bibliography{icdmrefs}

\end{document}